\documentclass{article}
\usepackage{ijcai23}

\usepackage{times}
\usepackage{soul}
\usepackage{url}
\usepackage[hidelinks]{hyperref}
\usepackage[utf8]{inputenc}
\usepackage[small]{caption}
\usepackage{graphicx}
\usepackage{amsmath}
\usepackage{pifont}
\usepackage{amsthm}
\usepackage{booktabs}
\usepackage{multirow}
\usepackage{algorithm}
\usepackage[switch]{lineno}
\usepackage{amssymb,amsfonts}
\usepackage{pbox}
\usepackage{cleveref}
\usepackage{subcaption}

\usepackage{xcolor}
\newcommand{\xmark}{\ding{55}}%
\newcommand{\cmark}{\ding{51}}

\definecolor{navyblue}{rgb}{0.0, 0.0, 0.5}

\newcommand{\etal}{\textit{et al}.}

\title{ State-wise Safe Reinforcement Learning: A Survey}

\author{
Weiye Zhao\and
Tairan He\and
Rui Chen\and
Tianhao Wei\And
Changliu Liu
\affiliations
Robotics Institute, Carnegie Mellon University\
\emails
\{weiyezha, tairanh, ruic3, twei2, cliu6\}@andrew.cmu.edu
\thanks{This material is based upon work supported by the National Science Foundation under Grant No. 2144489.}
}

\begin{document}

\maketitle

\begin{abstract}
    Despite the tremendous success of Reinforcement Learning (RL) algorithms in simulation environments, applying RL to real-world applications still faces many challenges. A major concern is safety, in another word, constraint satisfaction.
    State-wise constraints are one of the most common constraints in real-world applications and one of the most challenging constraints in Safe RL. 
     Enforcing state-wise constraints is necessary and essential to many challenging tasks such as autonomous driving, robot manipulation. This paper provides a comprehensive review of existing approaches that address state-wise constraints in RL. 
     Under the framework of State-wise Constrained Markov Decision Process (SCMDP), we will discuss the 
     connections, differences, and trade-offs of existing approaches in terms of (i) safety guarantee and scalability, (ii) safety and reward performance, and (iii) safety after convergence and during training. We also summarize limitations of current methods and discuss potential future directions.
\end{abstract}

\section{Introduction}
Reinforcement learning (RL) has achieved remarkable progress in games and control tasks~\cite{vinyals2019grandmaster,brown2018superhuman,he2022reinforcement,zhao2019stochastic,zhao2022safety}. However, one major barrier that limits the application of RL algorithms to real-world problems is the lack of safety assurance. 
RL agents learn to make decisions based on a reward signal, which may violate safety constraints. 
For example, an RL agent controlling a self-driving car may be able to achieve a high reward by driving at high speeds~\cite{zhao2019approximation}; but that 
may increase the chance to collide with others.
In other words, RL agents may sometimes prioritize maximizing the reward over ensuring safety, which can lead to unsafe or even catastrophic outcomes~\cite{gu2022review,zhao2022provably,he2023hierarchical}. 

Safe RL topics are emerging in literature, which investigate approaches to provide safety guarantees during or after the training. Early attempts are made under the framework of constrained Markov Decision Process, 
where the majority of works enforce cumulative constraints or chance constraints~\cite{liu2021policy}.
But in real-world applications, many critical constraints are instantaneous and deterministic. Not following these constraints may lead to catastrophic failure of the task. For example, collision avoidance for autonomous cars is a state-wise instantaneous constraint. It does not rely on historical trajectories or random variables. The constraints only depend on the current state of the system. Another example appears in robot manipulation~\cite{zhao2020contact}, e.g., when a robot holds a glass. The robot can only release the glass when it is placed on a stable surface. This constraint is neither cumulative nor probabilistic. It deterministically depends on the state of the robot. Violating this constraint leads to an irreversible failure of the task. 

The methods that specifically address state-wise hard constraints are studied as state-wise safe RL, where the problem is formulated as State-wise Constrained Markov Decision Process (SCMDP)
~\cite{berkenkamp2017safembrl,wachi2018safe,chow2019lyapunov}. 
State-wise safe RL needs policies that comply with the state-specific hard constraints.
Even though state-wise safe RL is challenging, it is also a critical step towards applying RL to real-world applications. Therefore we believe state-wise safe RL deserves more attention and effort from the RL community.

This paper serves as the first review of state-wise safe RL. Existing works are divided into two categories based on whether safety is ensured during training:
\begin{itemize}
    \item State-wise safety after convergence. When a policy converges, it ensures state-wise safety. But the policy may be unsafe during training.
    \item State-wise safety during training. The policy derived from this class of methods always ensures constraint satisfaction, even during training.
\end{itemize}
We also make fine-grained classifications based on their structures or assumptions to discuss the connections.


The remaining of the paper is as follows: We first formulate the problem of state-wise safety in \Cref{sec: formulation}. Then we summarize State-of-The-Art methods in \Cref{sec: convergence} and \Cref{sec: training}. In the end, we discuss trade-offs of existing methods in \Cref{sec: discussion} and future directions in \Cref{sec: future}

\section{Problem Formulation}
\label{sec: formulation}
\subsection{Preliminaries}
An Markov Decision Process (MDP) is specified by a tuple $(\mathcal{S}, \mathcal{A}, \gamma, \mathcal{R}, P, \rho)$, where $\mathcal{S}$ is the state space, and $\mathcal{A}$ is the control space, $\mathcal{R}: \mathcal{S} \times \mathcal{A} \rightarrow \mathbb{R}$ is the reward function, $ 0 \leq \gamma < 1$ is the discount factor, $\rho : \mathcal{S} \rightarrow [0, 1]$ is the starting state distribution, and $P: \mathcal{S} \times \mathcal{A} \times \mathcal{S} \rightarrow [0,1]$ is the transition probability function (where $P(s'|s,a)$ is the probability of transitioning to state $s'$ given that the previous state was $s$ and the agent took action $a$ at state $s$). A stationary policy $\pi: \mathcal{S} \rightarrow \mathcal{P}(\mathcal{A})$ is a map from states to a probability distribution over actions, with $\pi(a|s)$ denoting the probability of selecting action $a$ in state $s$. We denote the set of all stationary policies by $\Pi$. Subsequently, we denote $\pi_\theta$ as the policy that is parameterized by the parameter $\theta$. 

The standard goal for MDP, is to learn a policy $\pi$ that maximizes a performance measure, $\mathcal{J}(\pi)$, which is computed via the discounted sum of reward:
\begin{align}
\label{eq: reward function}
    \mathcal{J}(\pi) = \mathbb{E}_{\tau \sim \pi}\left[\sum_{t=0}^\infty \gamma^t \mathcal{R}(s_t, a_t, s_{t+1})\right],
\end{align}
where $\tau = [s_0, a_0, s_1, \cdots]$, and $\tau \sim \pi$ is shorthand for that the distribution over trajectories depends on $\pi: s_0 \sim \mu, a_t \sim \pi(\cdot | s_t), s_{t+1} \sim P(\cdot|s_t, a_t)$.

\subsection{State-wise Constrained Markov Decision Process}

A constrained Markov Decision Process (CMDP) is an MDP augmented with constraints that restrict the set of allowable policies for the MDP. Specifically, CMDP introduces a set of cost functions, $C_1, C_2, \cdots, C_m$, where $C_i : \mathcal{S} \times \mathcal{A} \times \mathcal{S} \rightarrow \mathbb{R}$ maps the state action transition tuple into a cost value. Similar to \eqref{eq: reward function}, we denote $\mathcal{J}_{C_i}(\pi) = \mathbb{E}_{\tau \sim \pi}[\sum_{t=0}^\infty \gamma^t \mathcal{C}_i(s_t, a_t, s_{t+1})]$ as the cost measure for policy $\pi$ with respect to cost function $C_i$. Hence, the set of feasible stationary policies for CMDP is then defined as follows, where $d_i \in \mathbb{R}$:
\begin{align}\label{eq: cumulative}
    \Pi_{C} = \{ \pi \in \Pi \big | ~\forall i, \mathcal{J}_{C_i}(\pi) \leq d_i\}.
\end{align}

In CMDP, the objective is to select a feasible stationary policy $\pi_\theta$ that maximizes the performance measure:
\begin{align}
\label{eq: original cdmp}
    \max_\theta \mathcal{J}(\pi_\theta), \text{ s.t. } \pi_\theta \in \Pi_{C}.
\end{align}

In this paper, we are specifically interested in a special type of CMDP where the safety specification is to persistently satisfy a hard cost constraint at every step, which is called State-wise Constrained Markov Decision Process (SCMDP). Similar to CMDP, SCMDP also uses the set of cost functions, $C_1, C_2, \cdots, C_m$, to evaluate the instantaneous cost for state action transition tuple. SCMDP requires the cost for every state action transition satisfies a hard constraint. Hence, 
the set of feasible stationary policies for SCMDP is defined as
\begin{align}
\label{eq:scmdp}
    \bar{\Pi}_{C} = \{ \pi \in \Pi \big | ~\forall (s_t, a_t, s_{t+1}) \sim \tau,~\forall i,~C_i(s_t, a_t, s_{t+1}) \leq w_i\}
\end{align}
where $\tau \sim \pi$ and $w_i \sim \mathbb{R}$. Then, The objective for SCMDP is to find a feasible stationary policy from $\bar{\Pi}_{C}$ that maximizes the performance measure. Formally, 
\begin{align}
\label{eq: fundamental problem}
    \max_\theta \mathcal{J}(\pi_\theta), \text{ s.t. } \pi_\theta \in \bar{\Pi}_{C}.
\end{align}
\vspace{-15pt}

\begin{figure}
     \centering
     \begin{subfigure}[b]{0.3\linewidth}
         \centering
         \includegraphics[width=\linewidth]{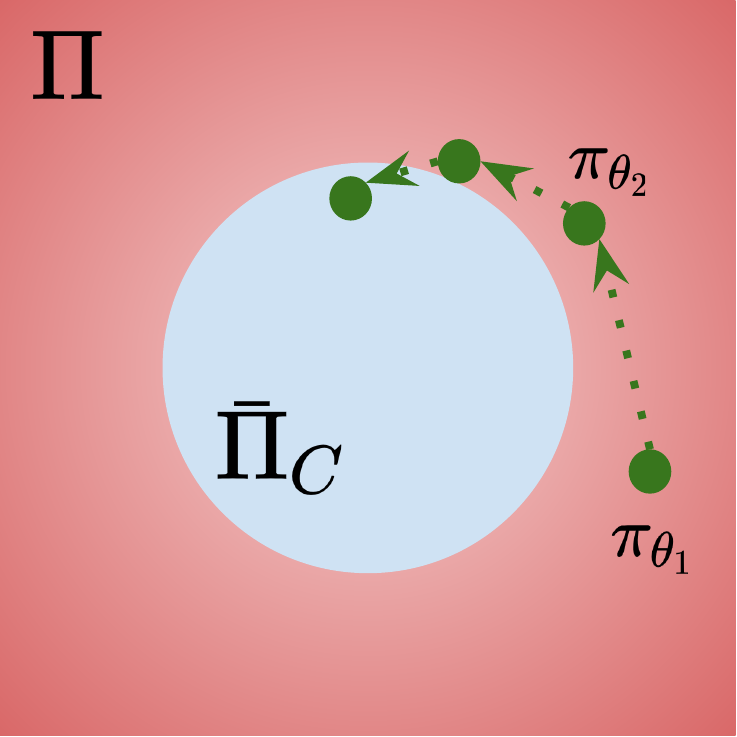}
         \caption{}
         \label{fig: a}
     \end{subfigure}
     \hfill
     \begin{subfigure}[b]{0.3\linewidth}
         \centering
         \includegraphics[width=\linewidth]{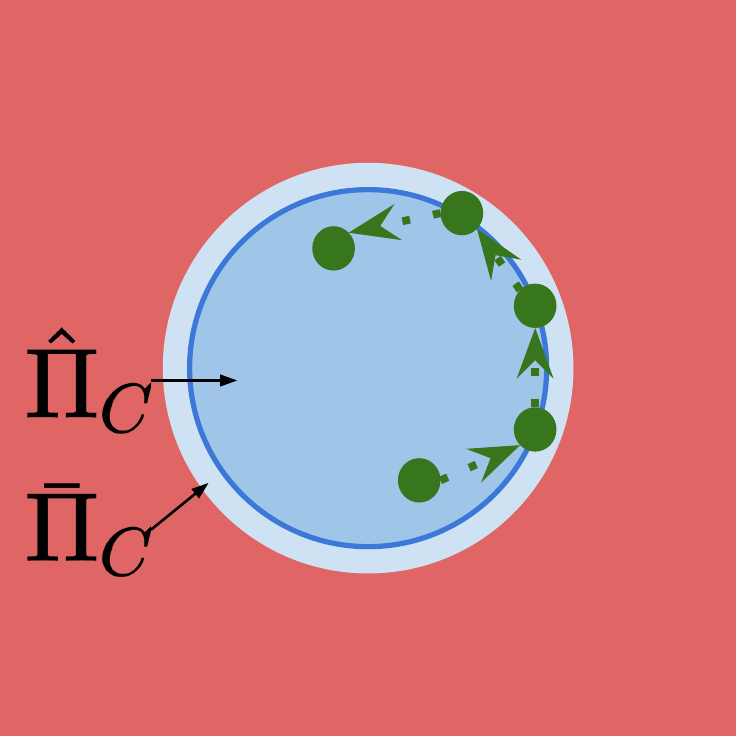}
         \caption{}
         \label{fig: b}
     \end{subfigure}
     \hfill
     \begin{subfigure}[b]{0.3\linewidth}
         \centering
         \includegraphics[width=\linewidth]{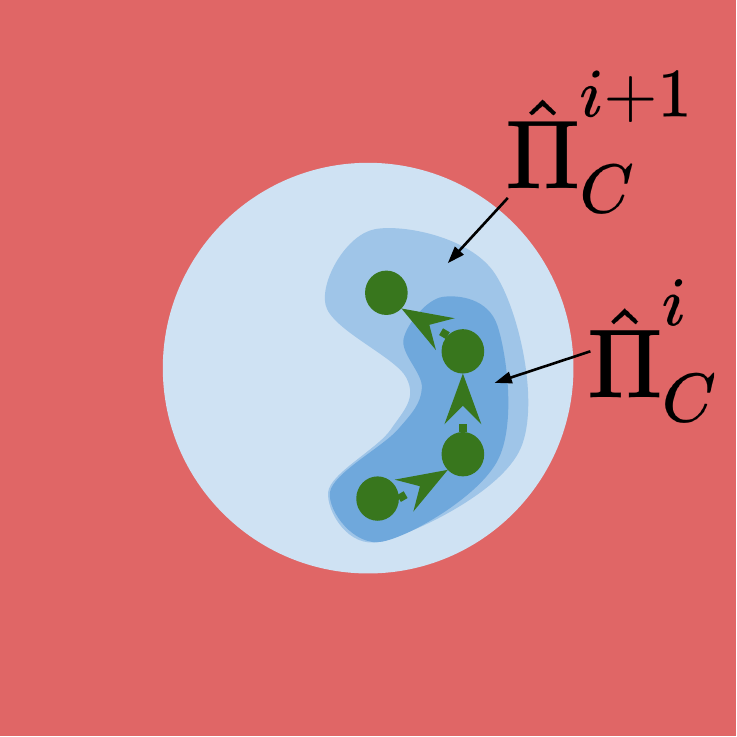}
         \caption{}
         \label{fig: c}
     \end{subfigure}
        \caption{Illustrations of 
        three different notions of state-wise safety
        (a) State-wise safety after convergence. This class of methods uses soft penalties to encourage policies to converge to the set of feasible policies ($\bar{\Pi}_C$). (b) State-wise safety during training by constraining policies in a subset of $\bar{\Pi}_C$. (c) State-wise safety during training by progressive safe exploration. }
        \label{fig: three graphs}
        \vspace{-10pt}
\end{figure}


    

\subsection{State-wise Safety After Convergence}

We denote the policy sequence during learning as $\pi_{\theta_1}, \pi_{\theta_2}, \cdots, \pi_{\theta_n}$
where $\theta_i$ denotes the policy parameters at optimization iteration $i$, and $n$ is the number of total iterations. State-wise safety after convergence refers to methods that have possibly infeasible policies during training and feasible converged stationary policies in the end. As shown in \cref{fig: a}, this class of methods uses soft penalties (typically called safe critic $Q_c(s,a)$) to guide the training of policies. The safe critic $Q_c(s,a)$ can be provided by humans or learned from the environment. This class of methods does not pose hard constraints on the exploration space, therefore
the policy learning can be made efficient. Nevertheless, the policy may be unsafe during training and the convergence to $\bar{\Pi}_C$ may be challenging in some cases.

\subsection{State-wise Safety During Training}

State-wise safety during training poses stricter requirements than state-wise safety after convergence because it requires all intermediate policies during training to be feasible. Namely,
$
    \forall i 
    \leq n, \pi_{\theta_i} \in \bar{\Pi}_C.
$
There are mainly two classes of methods to ensure state-wise safety during training. 

As shown in \cref{fig: b}, the first class defines a subset of $\bar{\Pi}_C: \hat{\Pi}_C$ (typically by control barrier functions). Whenever the policy goes out of the subset, it will be projected back. Therefore the policy is always guaranteed safe. However, $\hat{\Pi}_C$ requires a significant amount of prior knowledge about the environment, which may be lacking in many cases. And in general, these methods are more conservative than after-convergence methods because these methods can only explore a more limited state space. 
As shown in \cref{fig: c}, the second class of methods progressively explores the policy space from a safe initial policy. This class of methods maintains an up-to-date subset of $\bar{\Pi}_C$: $\hat{\Pi}_C^i$ by exploring and gathering information. The policy updates in $\hat{\Pi}_C^{i+1}$ only if there is enough information to confirm that the $\hat{\Pi}_C^{i+1} \subseteq \bar{\Pi}_C$. This class of methods requires less prior knowledge than the first class but is more conservative in exploration.

In the following sections, we will discuss the  representative works in both categories, i.e. state-wise safety after convergence and during training. \Cref{tab:main_table} summarizes the body of works that we consider in this survey. Based on the different policy solution types to \eqref{eq: fundamental problem}, we mainly classify the state-wise safe RL algorithms into (i) hierarchical agent, and (ii) end-to-end agent. Additionally, we compare the methods along the dimensions of state-wise safety theoretical guarantee and the associated assumptions.

\begin{table*}[t]
    \centering
    \resizebox{1\textwidth}{!}{
\begin{tabular}{ lllclll } 
\toprule
 & & Approach & 
Guarantee
& Assumption & $\bar{\Pi}_C$ evaluation & $\bar{\Pi}_C$ enforcement  \\
\midrule
\multirow{7}{0.5em}{\centering \rotatebox{90}{Asymptotic} } & \multirow{4}{0.5em}{\centering \rotatebox{90}{Structural}}  &  CSC~\cite{bharadhwaj2020conservative} & \xmark &  & Learned $C(s_t,a_t,s_{t+1})$ & Action sampling\\
&  & USL~\cite{zhang2022evaluating}  & \xmark & & Learned $C(s_t,a_t,s_{t+1})$ & Action optimization\\
& & LPG~\cite{chow2019lyapunov} & \xmark & & Learned $C(s_t,a_t,s_{t+1})$ & Action optimization \\
&  & RL-CBF~\cite{cheng2019end} & \cmark & White-box dynamics; well-calibrated GP & Learned $C(s_t,a_t,s_{t+1})$ & Action optimization\\
\cmidrule{2-7}
& \multirow{3}{0.5em}{\centering \rotatebox{90}{E2E}} & Lagrangian~\cite{bohez2019value} & \xmark & &  & Policy optimization\\
&  &   SSAC~\cite{ma2021learn} & \xmark & Pre-defined safety-oriented energy functions & Learned $V_C(s)$ & Policy optimization\\
&  &   AutoCost~\cite{he2023autocost} & \xmark & &  & Policy optimization\\
\midrule
\multirow{11}{0.5em}{\centering \rotatebox{90}{In-training}} & \multirow{8}{0.5em}{\centering \rotatebox{90}{Structural}} & RTS \cite{shao2021reachability} & \cmark & White-box system dynamics & Reachable set & Trajectory sampling\\ 
&  & SafeLayer~\cite{dalal2018safe}  & \xmark & Linear cost functions & Learned $C(s_t,a_t,s_{t+1})$ & Action optimization\\
&  & ShieldNN~\cite{ferlez2020shieldnn}  & \cmark  & Kinematic bicycle dynamics & Known $C(s_t,a_t,s_{t+1})$ & Action optimization\\
&  & HJB~\cite{fisac2018general} & \cmark  & White-box system dynamics & Known $C(s_t,a_t,s_{t+1})$ & Action optimization\\
&  & ISSA~\cite{zhao2021model} & \cmark & Black-box system dynamics (digital twins) & Forward simulation & Action sampling\\
&  & RecoveryRL~\cite{thananjeyan2021recovery} & \xmark &  & Learned $C(s_t,a_t,s_{t+1})$ & Policy optimization\\
&  & MIND~\cite{wei2022safe} & \cmark & Perfect learned system dynamics & Learned $C(s_t,a_t,s_{t+1})$ & Action optimization\\
&  & UAISSA~\cite{zhao2022probabilistic} & \cmark & Lipschitzness of system dynamics & Learned $C(s_t,a_t,s_{t+1})$ & Action optimization\\
\cmidrule{2-7}
& \multirow{3}{0.5em}{\centering \rotatebox{90}{E2E}} & Stability~\cite{berkenkamp2017safembrl} & \cmark &
 Lipschitzness of dynamics; default $\pi\in\bar{\Pi}_C$ & Known $C(s_t,a_t,s_{t+1})$ & Policy optimization\\
&  & SafeExpOpt-MDP~\cite{wachi2018safe} & \cmark & Black-box system dynamics & Learned $C(s_t,a_t,s_{t+1})$ & Action sampling \\
&  & SNO-MDP~\cite{wachi2020safe} &  \cmark & Black-box system dynamics & Learned $C(s_t,a_t,s_{t+1})$ & Action sampling\\
\bottomrule
\end{tabular}
}
    \caption{Overview of approaches for state-wise safe reinforcement learning, including (i) asymptotic and in-training safety, (ii) state-wise safety theoretical guarantee, (iii) assumptions, (iv) how to evalute $\bar{\Pi}_C$, and (v) how to enforce $\bar{\Pi}_C$. ``E2E'' stands for ``end-to-end''.}
    \label{tab:main_table}
    \vspace{-15pt}
\end{table*}

\section{State-wise Safety After Convergence}\label{sec: convergence}
\subsection{Hierarchical Agent}\label{sec:convergence_structure}
To enforce the state-wise safety constraint, a straightforward approach is that we evaluate the proposed action at every time step and project it into the 
safe action set, i.e. 
$\mathcal{A}_C(s_t) = \{a \sim \pi(s_t) \big| \pi \in \bar{\Pi}_C\}$, if it is unsafe. This results in a hierarchical safe policy, where (i) the upper layer is a reinforcement learning policy ($\pi_\theta$) that generates task-oriented action to maximize the cumulative reward $\mathcal{J}$, and (ii) the lower layer is a safety monitor that performs this safe action correction, such that this hierarchical policy belongs to $\bar{\Pi}_C$. To realize the safety monitor, the key issue is to learn a safety critic $Q_C(s_t,a_t)$ that 
estimates the chance to persistently satisfy $C(s_t, a_t, s_{t+1})$
from \eqref{eq:scmdp}, so that we can filter actions to ensure the hierarchical policy satisfies \eqref{eq:scmdp}. The major approaches for filtering can be divided into (i) directly sampling, and (ii) optimization-based projection.

With this inspiration, Conservative Safety Critics (CSC)~\cite{bharadhwaj2020conservative} is proposed to ensure the safe exploration during policy training. Their main intuition is to train a safe critic $Q_C(s,a)$ together with policy $\pi_\theta$ that provides a conservative estimate of the likelihood to be unsafe given a state action pair.
Here the conservative estimation means that the critic will overestimate the future cost led by the state and action pair. The safety critic will be used to monitor the future safety violation given proposed action from the current policy at every time step during policy evaluation. If the safety violation exceeds the predefined safety violation threshold, new actions will be re-sampled from the policy until the safety critic agrees the proposed action is safe. Mathematically, we 
\begin{align}
    \text{sample } a \sim \pi_\theta(s), \text{until } Q_C(s,a) \leq w~. 
\end{align}

Furthermore, the 
optimization of $Q_C$ and $\pi_\theta$
builds upon Constrained Policy Optimization (CPO)~\cite{achiam2017constrained}, which restricts the updated policy to stay close to its predecessor via trust-region method. Subsequently, 
the hierarchical policy will only generate actions that are safe w.r.t. the critic, but also inherit the theoretical guarantee on worst case cost violation from CPO via trust-region methods. On the other hand, CSC also inherits the limited generality and costly computation from CPO, since the policy optimization requires conjugate gradients for approximation of Fisher Information Matrix~\cite{achiam2017constrained}. Addtionally, imperfect cost function $Q_C(s,a)$ hurts the safety guarantee.

Although it is straightforward to directly sample the policy for a safe action as done in CSC~\cite{bharadhwaj2020conservative}, the sampling procedure is time consuming and may be detrimental to reward performance $\mathcal{J}(\pi_\theta)$.
A better strategy is to optimize the safe control so that it satisfies the constraint while stay close to the reference. And this optimization can be solved either by direct gradient descent or quadratic programming.
Under these types of approaches, Unrolling Safety Layer (USL)~\cite{zhang2022evaluating} is proposed to project the reference action into safe action via gradient-based correction. USL framework is comprised of (i) a pre-posed policy network that outputs task-oriented actions, and (ii) a post-posed USL to correct reference action such that state-wise safety constraints are satisfied. Specifically, USL iteratively updates a cost $Q_C(s,a)$ function (safety critic) with the samples collected during training, then USL performs gradient descent on the reference action
\begin{align}
    a_{new} = a_{old} - \frac{\eta}{\mathcal{Z}} \cdot \frac{\partial}{\partial a_{old}}[Q_C(s,a_{old}) - w]
\end{align}
until resulting cost satisfies the state-wise safety constraint, where $\eta$ is the step size, $\mathcal{Z}$ is a normalization fator that rescales the gradient. Emiprical results show that USL achieves state-wise zero safety violation after policy convergence in a variety of challenging safe RL benchmarks; this method, however, requires the perfect cost function $Q_C(s,a)$ approximation in order to formally guarantee state-wise safety, which is hard to prove.

Following the similar idea that safe action should be projected near task-oriented action, Lyapunov-based Policy Gradient (LPG) ~\cite{chow2019lyapunov} is also proposed to project action to the safe action set induced by linearized Lyapunov constraints. LPG builds upon the Laypunov-based approach to solve CMDPs~\cite{chow2018lyapunov}, where they first use LP-based algorithm to
construct Lyapunov functions with respect to generic CMDP constraints, then add a Lyapunov safety layer to safeguard any learning policy. During policy training, they use value iteration to estimate a state-action Lyapunov function $Q_C(s,a)$, then optimize for safe actions to satisfy Lyapunov constraint at every time step. Specifically, they linearize the Lyapunov constraint with its first-order Taylor series and define the objective to minimize the difference between safe action and reference action
\begin{align}
    &\min_{a} \frac{1}{2}\|a - a_{old}\|^2\\ \nonumber
    &~~~\text{s.t.} ~~ Q_C(a_{old}) + \nabla_{a}Q_C(s,a)|_{a = a_{old}} (a - a_{old}) \leq w.
\end{align}

Then, safe action can be optimized via Quadratic Programming (QP). However, the linearization approximation error is non-negligible for most Lyapunov constraints, and Lyapunov function approximation is usually imperfect, both facts hurt the state-wise safety guarantee.

Control Barrier Function (CBF) is famous for constraining the action to persistently satisfy state-wise safety constraints with known system dynamics. This motivates Richard \etal~\cite{cheng2019end} to propose a general safe RL framework, which combines control barrier function (CBF) based controllers with RL algorithms to guarantee safety and guide the learning process by constraining the set of explorable policies. The proposed RL-CBF framework first defines an energy function $\tilde{h}(s)$, so that $\forall t, C(s_t, a_t, s_{t+1}) \leq w$ is guaranteed via ensuring $\tilde{h}(s_t) \geq 0$ for all time step. Then RL-CBF assumes a known nominal control-affine dynamics model, i.e. $s_{t+1} = \tilde{f}(s_t) + \tilde{g}(s_t)a_t$ and adopts GP to model the remaining unknown system dynamics and uncertainties $\tilde{d}(s_t)$. Lastly, CBF can project the reference control into safe control via quadratic programming
\begin{align}
    &\min_{a} \frac{1}{2}\|a - a_{old}\|^2\\ \nonumber
    &~~~\text{s.t.} ~~ \tilde{h}(\tilde{f}(s_t) + \tilde{g}(s_t)a + \tilde{d}(s_t)) + (\tilde{\eta} - 1) \tilde{h}(s_t) \geq 0
\end{align}
where $\tilde{\eta}$ adjust how strongly the barrier function ``ensures" safety. During policy training, more state action transition data will be used to update the GP dynamics model. However, the safety guarantee of their CBF-based controller strongly relies on (i) the known affine system dynamics assumption, and (ii) the learnt GP model is well-calibrated to provide a high probability confidence interval for uncertain dynamics, which restricts its application to general nonlinear and high dimensional dynamics systems.

\subsection{End-to-End Agent}
End-to-End agent solves the constrained learning problem as a whole without any extra safety layer modules. The challenge is how to ensure the characterization of $\mathcal{J}_c$ enable policy to converge to the set \eqref{eq:scmdp}.
The usage of end-to-end learning schemes for state-wise safety after convergence has been explored in many recent works. 
Some safe RL works~\cite{liang2018accelerated,tessler2018reward} solve a primal-dual optimization problem to satisfy the CMDP constraint in expectation as shown in \eqref{eq: lagrangian_rl}.
\begin{equation}
\label{eq: lagrangian_rl}
\max_{\theta} \min_{\lambda \geq 0} \mathcal{L}(\pi_\theta, \lambda) = \mathcal{J}(\pi_\theta) -\sum_{i}\lambda_i (\mathcal{J}_{C_i}(\pi_\theta) - d)
\end{equation}
These methods are also applied to the setting of state-wise safety by augmenting the reward with the sum of the constraint penalty weighted by the Lagrangian multiplier~\cite{bohez2019value}. Mathematically, the augmented reward is defined as
\begin{equation}
\label{eq: lagrangian_rl_statewise}
\mathcal{R}(s_t, a_t, s_{t+1}) - \sum_i \lambda_i (C_i(s_t, a_t, s_{t+1}) - w_i)
\end{equation}
However, the optimization in \eqref{eq: lagrangian_rl_statewise} is hard in practice because the optimization updates for every RL step.
SSAC~\cite{ma2021learn} confines the Lagrangian-based policy update using safety-oriented energy functions. Intuitively, SSAC uses the safety-oriented energy function transition as the new constraint objective for policy updates, which enables SSAC to identify the dangerous actions prior to taking them and further obtain a zero constraint-violation policy after convergence. 
Note that SSAC is different from standard Lagrangian safe RL methods since SSAC uses an additional neural network $\lambda_\zeta(s)$ to approximate the state-wise Lagrange multiplier. Mathematically, SSAC solves
\begin{equation}
\label{ssac}
\max_{\theta} \min_{\zeta} \mathcal{L}(\pi_\theta, \lambda_\zeta) = \mathcal{J}(\pi_\theta) -\sum_{i}\sum_{s\sim\pi_\theta}\lambda_\zeta(s) (V^\pi_{C_i}(s) - d)
\end{equation}
where $V^\pi_{C_i}(s) = \mathbb{E}_{\pi}[\sum_{k=0}^{\infty} \gamma^k C_{i}(s_{t+k})|s_t = s]$
FAC-SIS~\cite{ma2022joint} formulates a loss function to optimize the safety certificate parameters by minimizing the occurrence of energy increase, relieving SSAC from the requirement of a perfect safety-oriented energy function.
To solve the problem of current standard safe RL algorithms (e.g., CPO and Lagrangian methods) failing to achieve zero-violation performance even when the cost threshold is set as zero, \citeauthor{he2023autocost} points out that one of the 
main bottlenecks to achieving zero violations lies in the lack of approaches to properly characterize the cost function in \eqref{eq:scmdp}. To address this, 
\citeauthor{he2023autocost} proposes AutoCost to find an appropriate cost function automatically using evolutionary search over the space of cost functions as parameterized by a simple neural network. The results suggest that the evolved cost functions enable agents to be both performant and not violate safety constraints at all after convergence.

\vspace{-5pt}

\section{State-wise Safety During Training}\label{sec: training}
\subsection{Hierarchical Agent}\label{sec:struct_during_training}
To satisfy state-wise safe constraints, a natural approach is to correct action at each step by projecting actions onto a feasible action such that the cost constraints are satisfied at every step. This can be done by 
constructing a hierarchical safe agent with upper layer generating reference actions, and lower layer performing the projection. Then, the solution towards \eqref{eq: fundamental problem} is a hierarchical-structured policy. Comparing with the hierarchical policy in \cref{sec: convergence}, the lower layer in this section usually has prior knowledge about $C(s_t, a_t, s_{t+1})$ via system dynamics assumption, hence is more capable of providing state-wise safety guarantee during policy training.

With the knowledge assumptions regarding the system dynamics, the existing hierarchical safe reinforcement learning solution is divided into three categories: 1) white-box dynamics, 2) black-box dynamics, and 3) learned dynamics. 

\subsubsection{White-Box Dynamics}
State-wise safety guarantee is extensively studied in the safe control related methods~\cite{noren2021safe,wei2022persistently}, such as control barrier functions, e.g. (CBF)~\cite{ames2014control}, safe set algorithms (SSA)~\cite{liu2014control}, which usually require the system dynamics to be explicitly known.
With the knowledge of analytical or explicit system dynamics model (white-box dynamics), the hierarchical-structured safe policy could leverage safe control methods to safeguard the learning policy, such that state-wise safety can be guaranteed. However, all these methods characterize the set of the stationary policies $\bar{\Pi}_C$ manually, which is hard to scalable.



ShieldNN is proposed ~\cite{ferlez2020shieldnn} to leverage Barrier Function (BF) to design a safety filter neural network with safety guarantees. ShieldNN adopts three major steps: 1) design a candidate barrier function, 2) verify the existence of safe controls, and 3) design a safety filter. The safety filter overrides any unsafe controls with the output from safe neural network controller. However, ShieldNN is specially designed for an environment with the kinematic bicycle model (KBM) \cite{kong2015kinematic} as the dynamical model, which is hard to generalize to other problem scopes. \cite{chen2021safe} consider a similar structure for a cluttered 2D environment. But they also use the safety filter to provide demonstration for the RL algorithm to improve sample efficiency.

~\cite{fisac2018general} also propose a general safety framework based on Hamilton-Jacobi reachability methods that can work in conjunction with an arbitrary learning algorithm. However, these methods still exploit approximate knowledge of the explicit form of system dynamics to guarantee state-wise safety constraints satisfaction.

~\cite{shao2021reachability} proposes a reachability-based trajectory safeguard (RTS) safety layer to ensure the system persistently satisfy safety constraint during training and operation. Specifically, RTS allows the RL agent to select high level trajectory plan instead of the direct control inputs, where the state-wise safety is guaranteed via ensuring the only safe plans are selected. To distinguish the safe and unsafe trajectory plans, RTS precomputes the forward reachable set of the robot when it tracks the parameterized trajectory. During the run time, robot will select safe plan based on the precomputed forward reachable set. However, the exact computation of forward reachable set is conditioned on the explicit knowledge system dynamics.


\subsubsection{Black-Box Dynamics}
To design a safety filter that projects the reference action into the feasible space, the critical step is to identify the feasibility of a certain action, i.e. whether the action is considered safe or unsafe. Subsequently, the state-wise safety is guaranteed via generating action that is considered to be safe at every step. Follow this inspiration, as long as the instantaneous cost for a state-action transition tuple can be evaluated, i.e. black-box dynamics model, safe and unsafe actions can be discriminated. Therefore, even without the knowledge of explicit and analytical form of system dynamics, it is possible to design a hierarchical-structured safe policy. Specifically, the hierarchical safe policy can leverage black-box model to recognize and filter the unsafe actions from RL policies.

Implicit safe set algorithm (ISSA)~\cite{zhao2021model} constructs a hierarchical-structured safe policy. The upper layer is an unconstrained RL algorithm PPO~\cite{schulman2017proximal} generating task-oriented reference action. The lower layer is safety layer that filters out unsafe control actions by projecting them to a set of safe control. ISSA first defines an energy function safety index $\phi(s)$, such that $\bar{\Pi}_C$ corresponds to all policies that ensures $\phi(s) \leq 0$ for all time step. The construction of $\phi(s)$ does not require explicit knowledge of system dynamics. The only thing need is the kinematic limits -- hence this can significantly address the scalability issue with the white-box models.
With the black-box dynamics model $s_{t+1} = f(s_t, a_t)$, ISSA ensures state-wise safety guarantee via solving safe control at every time step
\begin{align}
\label{eq: issa}
    &\min_{a} \frac{1}{2}\|a - a_{old}\|^2 \quad
    &~~~\text{s.t.} ~~ \phi(f(s_t, a)) \leq 0
\end{align}
where \eqref{eq: issa} is solved via multi-directional line search, and $\phi(f(s_t, a))$ can be checked via one-step simulation through black-box dynamics model, e.g. a digital twin simulator.

\subsubsection{Learned Dynamics}


To relieve the requirements of explicit system dynamics, the last category of methods provide safety guarantees based on learned models from offline datasets. ~\cite{dalal2018safe} propose to learn the system dynamics directly from offline collected data, then add a safety layer that analytically solves a control correction formulation per each state to ensure every state is safe. However, the closed-form solution relies on the assumptions of (i) linearized dynamics model, and (ii) linearized half-space shaped safe action set,
which does not hold for most of the real-world dynamics systems.

~\cite{thananjeyan2021recovery} propose a Recovery RL algorithm, which uses the offline collected dataset to train a safety critic module that quantifies the the likelihood of certain action that leads to safety violation in the near future. In the meanwhile, a recovery policy that minimizes the risk of constraint violation is also pretrained based on the offline dataset. During online training, the task policy optimizes the task reward, and recovery policy action will override the task policy action when it is considered unsafe by the safety critic. However, since there's no guarantee that pretrained safety critic can correctly identify the  unsafe action, the resulting recovery policy is biased and safety violation is inevitable. 


To apply CBF for end-to-end neural network dynamic models, Wei \etal~\cite{wei2022safe} use an evolutionary strategy to design CBF-based safety filters. They also propose a mix-integer programming method to find optimal safe actions given the CBF, which is called Mixed Integer for Neural network Dynamic models (MIND). MIND can be incorporated to safeguard arbitrary RL algorithms. However, their methods have difficulty scaling to high dimensions because the data required to design such a CBF filter grows exponentially with the dimensions. Additionally, the state-wise safety guarantee comes with a assumption that learnt neural network dynamics exactly capture the ground-truth system dynamics. 


To achieve zero safety violation during training for general dynamics system,
~\cite{zhao2022probabilistic} propose Uncertainty-Aware Implicit Safe Set Algorithm (UAISSA) to probabilistically safeguard robot policy learning using energy-function-based safe control with a GP dynamics model learned on an offline dataset, and the system dynamics is a general nonlinear function. 
To achieve the zero safety violation during training with GP dynamics model, in the offline stage, they first show how to construct the dataset for system dynamics learning and how to design the associated energy function (called \textit{safety index})~\cite{liu2014control} so that there always exists a feasible safe control for all possible states under control limits. 
Secondly, they design a safety layer to project the reference control the nominal learning policy into a set of safe control via solving the same optimization problem as \eqref{eq: issa}, where one step forward simulation is conducted with the learned GP dynamics model. With the guarantee of nonempty set of safe control for all possible states, the method is able to achieve zero constraint violation during policy learning. One drawback, however, is that the dynamics learning dataset for GP requires grid sampling of the state space, which makes the method hard to scale to higher dimensional systems.

\subsection{End-to-End Agent}

Besides hierarchical safe agents, it is also possible to construct end-to-end agents that satisfy state-wise training time safety.
In this section, we focus on agents without prior knowledge of the environment, i.e., how to project actions to 
an uncertain safe action set, considering the uncertainty of learned dynamics.
Those with such prior will yield formulations similar to  constrained optimizations, which essentially apply methods discussed in \cref{sec:convergence_structure} at training time.
There could be another type of approach that initially guards the end-to-end agent with a safety monitor, and gradually remove the monitor as learning progresses.
Approaches of such kind are out of the scope of this survey.

One representative method of interest proposed by \cite{berkenkamp2017safembrl} learns a single policy that directly generates actions with high-probability safety guarantees at every step during training.
Inspired by Lyapunov stability theories \cite{khalil2002nonlinear}, the method defines safe policies to be those which keep the system within a subset of the state space, or region of attraction (ROA), if the system starts within that subset.
The ROA is defined as the level set $\mathcal{V}(c)$ of a Lyapunov energy function $v:\mathcal{S}\mapsto \mathbb{R}$, e.g., $\mathcal{V}(c)=\{s\in\mathcal{S} \backslash \{\mathbf{0}\} \mid v(s)\leq c \}$.
The objective is to simultaneously learn the unknown system dynamics and acquire high-performance control policies while ensuring that the state never leaves the ROA.
The goal is achieved by maintaining a safe policy $\pi$ with an estimate of the dynamics $f$ modeled by Gaussian processes \cite{williams2006gaussian}, and iteratively updating both while taking safe actions only.
In each iteration, the algorithm first finds the largest ROA allowed by the current policy $\pi$.
Then, it drives the system to the state-action pair that the current dynamics model $f$ is most uncertain about, and updates $f$ using the measurement there.
Finally, the policy $\pi$ is optimized to maximize rewards.
The policy optimization with ROA estimation can be formulated as
\begin{equation}\label{eq:stable_obj}
    \pi,c = \underset{\pi,c}{\mathrm{argmax}}~c,~\mathrm{s.t.}~\forall s\in \mathcal{V}(c),~\pi(s)~\mathrm{reduces}~v.
\end{equation}
Starting with an initially safe policy and a conservative estimate of the ROA, the above procedure progressively gathers safe and informative data points to improve the dynamics model, which in turn expands the estimated ROA.
The procedure is proven to achieve full exploration within reachable parts of the state space.
This work finally transfers the theoretical results to a practical algorithm and shows that it can optimize a neural network policy on an inverted pendulum simulation without letting the pendulum ever fall.
One drawback of this work is the difficulty of obtaining a proper Lyapunov function for defining the ROA.
The method is also prone to non-stationary system dynamics since past observations on state transitions might turn invalid for GP regression.

SafeExpOpt-MDP~\cite{wachi2018safe} is another end-to-end approach that satisfies safety during training.
This approach balances a three-way trade-off of exploring the unknown safety function, exploring the reward function, and maximizing (exploiting) rewards.
The state-wise safety is guaranteed by only navigating to a subset $\mathcal{S}^\mathrm{safe}$ of the state space where the states are (a) safe even considering worst-case prediction error, (b) one-step reachable from a safe state, and (c) returnable to a safe state.
In \cite{berkenkamp2017safembrl}, similar safety guarantees are ensured as the forward invariance within a safe region.
The exploration and exploitation of learned rewards are encoded by an objective that linearly interpolates between an optimistic value function $\hat{\mathcal{J}}$ and a pessimistic one $\bar{\mathcal{J}}$ as
\begin{equation}
    \pi = \underset{s\in\mathcal{S}^\mathrm{safe}}{\mathrm{argmax}}~(1-\eta)\bar{\mathcal{J}} + \eta\hat{\mathcal{J}}
\end{equation}
where $\eta\in[0,1]$.
Notably, SafeExpOpt-MDP is not near-optimal in terms of the cumulative reward.
As an improvement, SNO-MDP~\cite{wachi2020safe} achieves near-optimality in rewards under regularity assumptions while preserving safety guarantees.
SNO-MDP operates in two steps.
Firstly, the agent expands a pessimistic safe region while guaranteeing safety.
Then, it explores and exploits the reward in the safe region certified by the first step.
Incorporating results from probably approximately correct Markov decision process (PAC-MDP) \cite{strehl2006pac}, it is shown that the acquired policy is $\epsilon$-close to the optimal policy in the safety-constrained MDP.
The key condition that allows near-optimality is the optimistic estimation $U$ of the reward function with upper confidence bound \cite{chowdhury2017kernelized} as $U(s)=\mu^r(s)+\alpha^{1/2}\sigma^r(s)$, where $(\mu^r,\sigma^r)$ are the mean and variance of the estimated reward distribution and $\alpha$ is the scaling factor.

\section{Discussion}\label{sec: discussion}
In this section, we discuss two trade-offs to consider when designing a state-wise safe RL algorithm.
\subsubsection{Trade-Off Between Guarantee and Scalability}
As shown in \Cref{tab:main_table}, every theoretical safety guarantee comes with a price of making assumptions about the dynamics system or access to some prior knowledge. 
Most assumptions are made about dynamics including (i) analytical form or Lipschitz continuity of dynamics; (ii) perfect simulation of the dynamics; (iii) offline datasets of the dynamics transition. 
With strong assumptions (e.g., white-box/black-box dynamics), deriving a solid safety guarantee can be straightforward for specifically limited systems~\cite{ferlez2020shieldnn,fisac2018general,zhao2021model}.
With weaker assumptions (e.g., datasets of dynamics), the corresponding guarantee also becomes weaker (e.g., probabilistic safety guarantee~\cite{cheng2019end,zhao2022probabilistic}. )
However, some assumptions are hard to meet in real-world problems, which makes it hard to scale to other problem scopes. For example, it is hard to derive theory from complicated dynamics like snake robots and humanoid robots.
But as we make weaker and fewer assumptions, the subsequential state-wise safe RL algorithms also enjoy better scalability to different dynamics systems~\cite{bohez2019value,ma2021learn,he2023autocost}, which in turn comes with a price of losing state-wise safety theoretical guarantee.


\subsubsection{Trade-Off Between Asymptotic and In-Training Safety}
Asymptotic safety (state-wise safety after convergence) is often desired in situations where the agent's actions have a low probability of causing harm in the early stages of training, but the risk increases as the agent becomes more proficient. The main disadvantage of asymptotic safety is usually losing the theoretical safety guarantee, whereas the main advantages are (i) less restrictive assumptions and easy to scale, (ii) agents can explore broader state space, e.g. unsafe states, which is beneficial for improving the reward performance.

On the other hand, strict safety from the beginning of training can be more difficult to implement, as it often requires a significant amount of prior knowledge about the environment or the task to be able to define safe actions. The main advantage of this approach is that it eliminates the risk of the agent causing harm during the training process.
Both safety desires have their own benefits and drawbacks, and the best approach relies on the application and the level of safety required.


\section{Summary and Future Prospects}\label{sec: future}
Recently, state-wise safe RL has become an inspiring domain, motivated by both scientific challenges and industrial demands.
We believe state-wise safe RL is a major milestone of applying RL to real-world applications like robotics.
This paper provides a brief review of a considerable amount of studies of state-wise safe RL. Interested readers can also refer to some other survey papers~\cite{liu2021policy,gu2022review,brunke2022safe} that focus on more general safe RL settings (e.g., risk-sensitive RL, expectation-constrained RL).

Despite the fast development and great success of state-wise safe RL algorithms, significant challenges in this field still remain.
We conclude this paper with three key challenges: assumptions on system dynamics, optimality in terms of cumulative rewards, and asymptotic safety guarantees for end-to-end approaches.


\subsubsection{Prior Knowledge on the System Dynamics}



Many methods in this survey rely on assumptions about system dynamics, limiting their practical applicability. The assumption of white-box dynamics necessitates explicit knowledge of dynamics equations, which is rarely available except for a few cases such as controlled environments with known physical models. The assumption of black-box dynamics requires access to the target system, which can be costly or risky due to high sample complexity. The common assumption of Lipschitz continuity is weaker but still challenging to verify in practice. Therefore, learning dynamics is considered the most promising approach for relaxing these assumptions, given that model mismatch issues can be mitigated. One direction is to detect model errors, such as deviations in system properties like Lipschitz constant, even after convergence, allowing for continual correction of the dynamics model.

\subsubsection{Optimality Regarding Cumulative Reward}
Another challenge is the optimality of acquired policies in terms of cumulative reward.
This challenge is caused by the inherent conflict between the need for exploration for learning optimal RL policies and the need for safety.
The challenge manifests itself differently for different approaches.
For hierarchical agents, the existence of safety layers effectively distorts the reward signal in an intractable fashion, making it hard for nominal policies to improve.
For end-to-end methods with in-training safety guarantees, the exploration of rewards is highly limited by the gradually expanding safe region, which also leads to difficulties of learning.

\subsubsection{Asymptotic Safety Guarantee for End-to-End Methods}

Obtaining state-wise safety guarantees remains challenging for end-to-end methods after convergence. However, achieving asymptotic safety, though weaker than in-training safety, is still highly valuable in practical scenarios where the risks of unsafe states before convergence can be contained. While state-wise safety guarantees may be absent, theoretical guarantees for discounted cumulative cost in problem \eqref{eq: original cdmp} exist. Notably, CPO is a representative method that provides worst-case cumulative safety violation guarantees. This guarantee is derived by manipulating cumulative cost over infinite-horizon trajectories, but similar derivations can be used to extend the worst-case safety constraint violation to state-wise costs.

\section*{Contribution Statement}
\textbf{Weiye Zhao}: Conceptualization, Investigation, Formal analysis, Writing (original draft). \textbf{Tairan He}: Investigation, Formal analysis, Writing (original draft). \textbf{Rui Chen}: Investigation, Formal analysis, Writing (original draft). \textbf{Tianhao Wei}: Investigation, Formal analysis, Writing (original draft). \textbf{Changliu Liu}: Conceptualization, Writing (review \& editing), Supervision, Project administration, Funding acquisition.
\bibliographystyle{named}
\bibliography{ijcai23}

\end{document}